\documentclass[letterpaper, 10 pt, conference]{ieeeconf}  

\IEEEoverridecommandlockouts                              

\usepackage{graphicx}
\usepackage{comment}
\usepackage{amsmath,amssymb} 
\usepackage{color}
\usepackage{siunitx} 
\usepackage{etoolbox}
\makeatletter
\patchcmd{\@makecaption}
{\scshape}
{}
{}
{}
\makeatletter
\patchcmd{\@makecaption}
{\\}
{:\ }
{}
{}
\makeatother

\usepackage{multirow}
\usepackage[dvipsnames,table,xcdraw]{xcolor} 

\def\secref#1{Sec.~\ref{#1}}
\def\figref#1{Fig.~\ref{#1}}
\def\tabref#1{Tab.~\ref{#1}}
\def\eqref#1{Eq.~(\ref{#1})}

\title{\LARGE \bf
SKD: Keypoint Detection for Point Clouds using Saliency Estimation
}

\author{Georgi Tinchev, Adrian Penate-Sanchez and Maurice Fallon
	\thanks{The authors are with the Oxford Robotics Institute, University of Oxford, United Kingdom. 
	\texttt{\{gtinchev,adrian,mfallon\}@robots.ox.ac.uk}}%
	\thanks{This work was supported by EPSRC RAIN and ORCA Robotics Hubs (EP/R026084/1 and EP/R026173/1 respectively) 
	and EU H2020 project Memory of Motion (MEMMO, project ID: 780684). M. Fallon is supported by a Royal Society University 
	Research Fellowship.}
}

\begin{document}

\maketitle
\thispagestyle{empty}
\pagestyle{empty}

\begin{abstract}
	We present SKD, a novel keypoint detector that uses saliency to determine the best candidates from a point cloud for 
tasks such as registration and reconstruction. The approach can be applied to any differentiable deep learning descriptor by 
using the gradients of that descriptor with respect to the 3D position of the input points as a measure of their saliency. The 
saliency is combined with the original descriptor and context information in a neural network, which is trained to learn robust 
keypoint candidates. The key intuition behind this approach is that keypoints are not extracted solely as a result of the geometry 
surrounding a point, 
but also take into account the descriptor's response. The approach was evaluated on two large LIDAR datasets - the Oxford 
RobotCar dataset and the KITTI dataset, 
where we obtain up to 50\% improvement over the state-of-the-art in both matchability and repeatability. When 
performing sparse matching with the keypoints computed by our method we achieve a higher inlier ratio and faster 
convergence.
\end{abstract}


\section{Introduction}
\label{sec:introduction}

A key task in robotics is the ability for a robot to localize itself in its surrounding environment. There are two prominent approaches - 
dense registration and sparse keypoint matching of points to a known map or object. Dense registration methods try to match every point 
to the map while sparse keypoint matching methods choose points with similar characteristics. Both methods have weaknesses: dense
registration methods suffer from occlusions or modifications in the environment while keypoint matching requires a descriptive environment 
with repeatable points. 

In this work, we are concerned about the challenges of sparse feature matching methods. Such methods are usually carried out 
in two stages: keypoint extraction, in which \emph{good} points are selected according to some criteria, and keypoint 
description, in which the points are characterized.  We present a novel method for keypoint extraction from point clouds that 
exploits the saliency of the 
point's descriptor to propose points with higher likelihood of being matched. This follows the recently proposed 
\textit{describe-then-detect} methodology~\cite{dusmanu2019d2}, where we focus mainly on the \textit{detect} part, assuming 
an already trained differentiable point cloud descriptor.



\begin{figure}[!t]
	\begin{center}
		\includegraphics[trim={0 6cm 0 0},clip,width=1.0\linewidth]{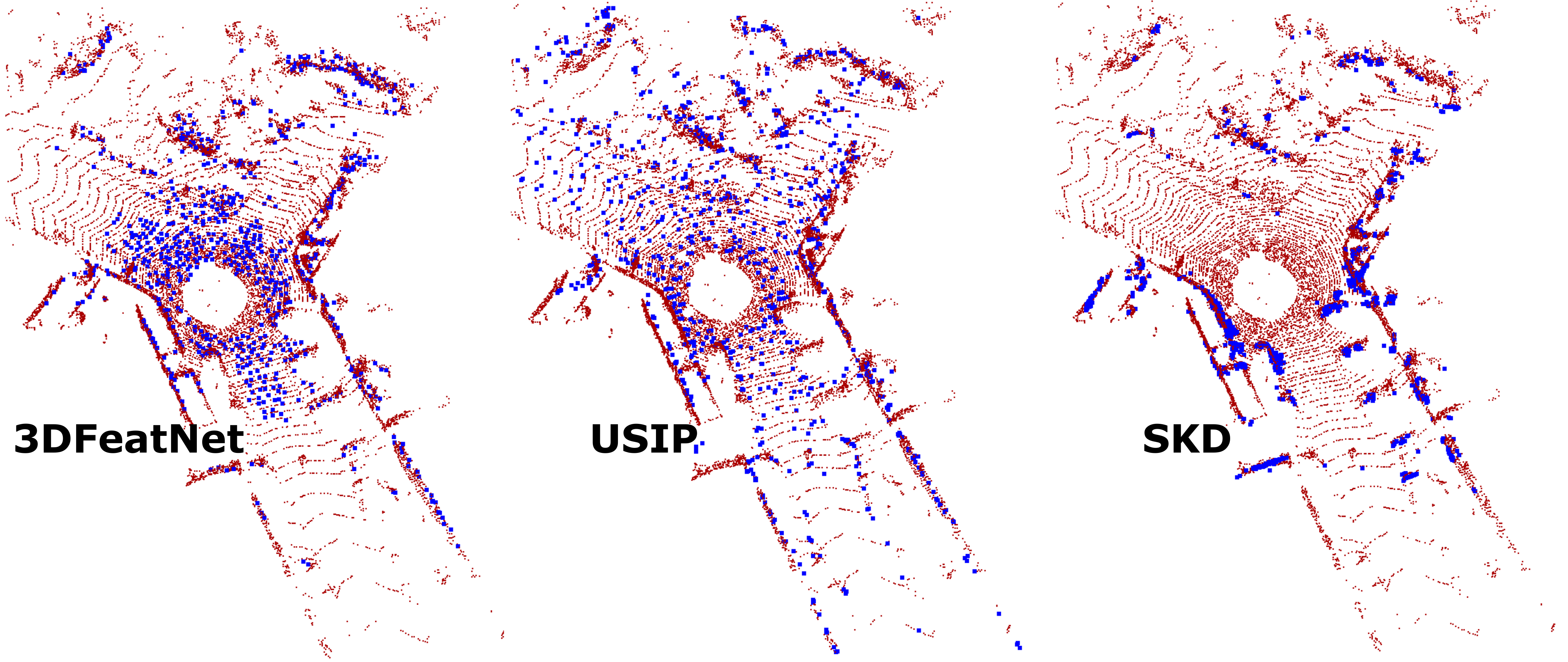}
	\end{center}
	\vspace{-5mm}
	\caption{Top-down view of a single point cloud (brown) and 1024 generated 
		keypoints (blue). Our method (Salient Keypoint Detector, SKD) was not implicitly trained to ignore the 
		ground, but	rather it uses the feature signal of a descriptor network and the context information to 
		determine informative areas in the environment - corners, edges and structure.
	}
	\label{fig:repeatability_visual}
	\vspace{-7mm}
\end{figure}

Keypoint extraction methods have focused on analyzing the appearance of a 
sensor stream (mainly images or point clouds) so as to reliably generate 
keypoints at the same location when faced with different 
observation of the same place or object; this is known as \textit{repeatability}. 
However, it presents some issues depending on the particular object or place 
being analyzed. For example, when extracting visual keypoints and descriptors 
from a chessboard, keypoint extraction can be performed reliably and those 
points will be \textit{repeatable} but the descriptors of those points will not 
be useful due to the ambiguity caused by the repeated pattern. The same can 
occur with point cloud data in which the windows of a building look alike. 
For this reason, recent research suggests that repeatability by itself is not 
sufficient for evaluating keypoint quality~\cite{r2d2,elf_ICCV19}. A more 
reliable metric is \textit{matchability}, as it measures how useful the 
combination of \emph{keypoint and feature} is. 

Our approach learns to produce points that can be reliably matched when given a 
pre-trained differentiable descriptor network. We leverage the idea presented in~\cite{elf_ICCV19} for image data, in which 
state-of-the-art keypoint detectors can be built by only looking at the feature 
response. We use the gradient response of the descriptor network with respect 
to the input points as 
a measure of \emph{saliency}, since it has been shown to produce promising 
results both in images and in point 
clouds~\cite{Simonyan14a,DB15a,pointCloudSal_ICCV19}. Further, we combine the 
saliency with the original descriptor and context information in a neural 
architecture, which is trained to produce keypoints that generate a higher 
percentage of inlier matches, shown in~\figref{fig:repeatability_visual}. The main 
contributions of our work are as follows:


\begin{itemize}

	\item The Salient Keypoint Detector (SKD): a novel method for extracting keypoints in point clouds based on the 
	gradient response of a differentiable descriptor and the geometry of the scene.
	\item An environment agnostic approach: by using the gradient response rather than the actual point cloud, our 
	approach is less vulnerable to biases in the training data. This makes our method robust when testing on different 
	environments and allows the model to be used without retraining for different datasets.
	\item State-of-the-art performance: Compared to the state-of-the-art~\cite{3dfeatnet,li2019usip}, our approach generates two 
	times more correct matches and achieves 40\% more relative repeatability. We have evaluated our algorithm on two of the largest 
	and most widely used LIDAR datasets - the Oxford RobotCar dataset~\cite{RobotCarDatasetIJRR} and the KITTI 
	dataset~\cite{kitti2013IJRR}. In total these datasets contain point clouds collected from more than 300 kilometers of 
	driving.
				
\end{itemize}

\section{Related Work}
\label{sec:related_work}

This paper focuses on reliably obtaining keypoints from
point clouds. This is a topic which is very closely related to
image keypoint extraction, so we will review the literature for both fields in this section. We will
also outline the relevant research on point cloud-based deep
learning models and saliency methods as these topics are central to the contributions of our approach.

\subsection{Image Keypoint Detectors}

Much of the recent work on keypoint and feature extraction from images employs deep learning architectures. 
In~\cite{Georgakis2018} a keypoint detector for depth images was presented. It used a Siamese approach by pairing two 
Faster-RCNN networks with a contrastive loss between them. Focusing on unsupervised 
keypoint learning,~\cite{deepmindNIPS19} presented an approach that generated reliable keypoints by learning from the temporal 
consistency of the network activations in short videos. A successful approach for obtaining good keypoints is to optimize how they
respond to the content of the image and not just the geometry. TILDE~\cite{VerdieYFL14} learned to make the points more 
reliable by understanding how changes in weather and lighting conditions modified the performance of a point.
In LF-Net~\cite{lf_net}, the keypoint detector and the feature descriptor are jointly learned by 
leveraging structure from motion (SfM) sequences to generate large amounts of training data. By understanding that 
the job of a keypoint is to increase the probability of descriptor matching, it focused on matching performance and 
at the time outperformed all previous approaches. Recently, ELF~\cite{elf_ICCV19} applied a simple approach by 
using the gradient response of deep learning features to produce keypoints.

\begin{figure*}[t!]
	\begin{center}
		\vspace{2mm}
		\includegraphics[trim={0 0.6cm 0cm 0.1cm},clip,width=0.88\linewidth]{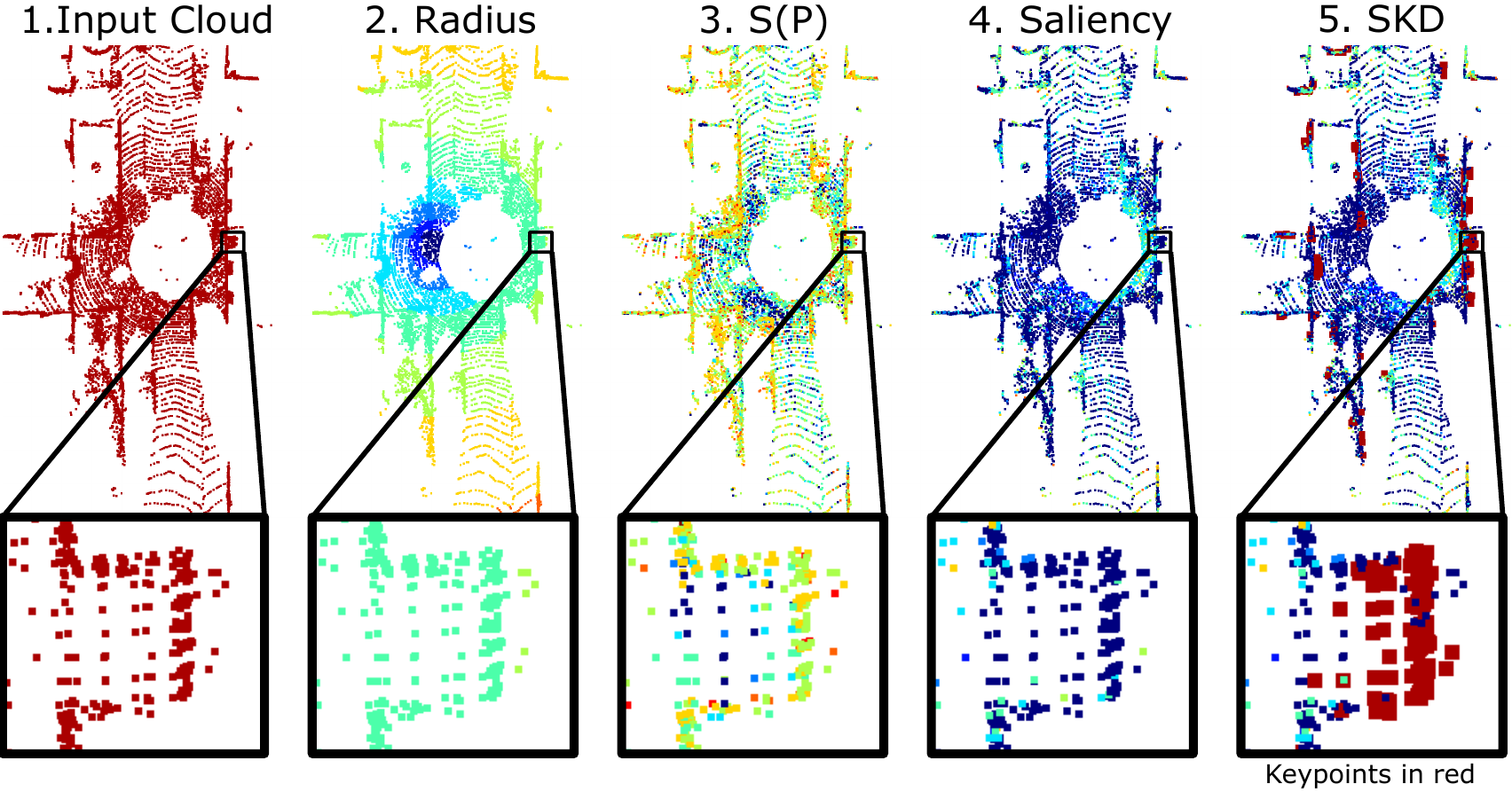}
	\end{center}
	\vspace{-4mm}
	\caption{\textbf{1.} Input point cloud $P$. \textbf{2.} Distances of each point ($r_i$) from the point cloud median - blue 
	indicates closer to the median point, red - further apart. \textbf{3.} The \textit{initial saliency}, 
		$S(P)$, based on the activations and gradients w.r.t. input point cloud (1.) - blue corresponds to negative 
		gradients, 
		red - highly positive. \textbf{4.} The \emph{saliency} score $s_i$ estimated from the radius (2.) 
		and the gradients (3.), blue - low score, red - high score. \textbf{5.} Selected keypoints (red, 
		enlarged) on top 
		of the saliency score. Our method picks keypoints based on the keypoint saliency score $s_i$, but 
		also leverages 
		context and the original descriptor to correctly pick out edges and corners negatively scored by 
		$s_i$.}
	\label{fig:approach_qualitative}
	\vspace{-6mm}
\end{figure*}

R2D2~\cite{r2d2} not only focused on repeatability for a keypoint, but also on the reliability score based on the descriptor's 
response. They jointly learned the keypoint detector and the discriminativeness of the descriptor. This is similar to our 
proposition to look at matchability of a keypoint in addition to its repeatability. SuperPoint~\cite{detone2018superpoint} 
jointly learned detection and description in a supervised task with multiple training stages on artificial images of 
basic structures. D2-Net~\cite{dusmanu2019d2} used a CNN with shared weights for description and 
detection. In contrast to previous approaches it used a methodology of \textit{describe-then-detect}. 
UnsuperPoint~\cite{christiansen2019unsuperpoint} improved the 
training methodology of SuperPoint by requiring only a single training round. They use a self-supervised detector and 
unsupervised descriptor to improve the repeatability of proposed keypoints. Their detector and descriptor also shared a common 
backbone. SIPs~\cite{cieslewski2019sips} introduced a metric to minimize the number of detected points with an iterative 
algorithm. Similar to us, they used the probability of a pixel being an inlier, but in 2D image data.

An alternative 
research direction was taken by GLAMpoints~\cite{truong2019glampoints} to apply reinforcement learning rewards to SIFT matches 
in order to maximize correct correspondences. Similarly, DISK~\cite{tyszkiewicz2020disk} leveraged 
cooperative reinforcement learning to match local features with a policy gradient.
In contrast to the above works, we do not jointly train a keypoint detector and descriptor, but instead focus only on the former 
and assume a given keypoint descriptor. Therefore, our methodology can be viewed as similar to \textit{describe-then-detect}, 
with a given differentiable \textit{describe} method. Also, we do not explicitly define a reward for each correct keypoint, nor use 
any filtering techniques.

\subsection{Point Cloud Keypoint Detectors}

PointNet~\cite{qi2016pointnet}
and PointNet++~\cite{qi2017pointnetplusplus} introduced new ways to efficiently
understand unstructured sets of points (point clouds). These works presented a novel layer for extracting
descriptions from sets of points by learning a symmetric function and thus making the layer invariant to 
point order. PointCNN~\cite{pointCNN_NIPS18} managed 
to increase the performance of the basic neurons by approximating a convolution using an MLP and by performing aggregation of spatial data 
within each neuron. PointNetVLAD~\cite{uypointnetvlad} presented a method to perform place recognition from point clouds. The point clouds 
were described using PointNet++ and NetVLAD trained with a metric learning loss to produce a feature vector. In contrast, we focus on 
detecting keypoints that aid descriptors such as~\cite{qi2016pointnet,qi2017pointnetplusplus,pointCNN_NIPS18}. Many recent papers 
leveraged PointNet layers in different applications~\cite{Xiang_2019_CVPR,pointCloudSal_ICCV19,pointFusion_CVPR18}.
In~\cite{pointCloudSal_ICCV19} the authors built a saliency map to 
understand the effect of each point on the final prediction. They 
tested the performance of their saliency score by performing point dropping 
operations to demonstrate performance better than a method based on 
the critical-subset theory~\cite{pointCloudSal_ICCV19}. We take inspiration from~\cite{pointCloudSal_ICCV19}, but we do not perform point 
shifting or point dropping operations. 

The most relevant methods to our work are 3DFeatNet~\cite{3dfeatnet} and USIP~\cite{li2019usip}.
3DFeatNet uses PointNet++ as a building block and learned a
detector and a descriptor by using a two stage network. Unfortunately, the approach did not explicitly focus on good 
performance of the keypoint extraction network. USIP aimed to mitigate this 
explicitly by minimizing a variation of the Chamfer loss that focused on matching points reliably. In contrast, 3DFeatNet used a 
triplet loss, that focused in obtaining a descriptive feature space, but that cannot manage matching
as it only focused on the descriptor. Compared to our work, 3DFeatNet~\cite{3dfeatnet} extracted attention for each point in the 
input, while we use the saliency scores obtained directly from the feature descriptor.

Recently, DH3D~\cite{dh3d} introduced a hierarchical Siamese network for joint learning of local description, keypoint score, and 
global descriptor. It uses a coarse-to-fine re-localization technique to first find submaps in the scene and then local features for 
accurate pose estimation. Similarly to 3DFeatNet, they used two-phase training to first train the descriptor and then the 
detector. D3Feat~\cite{d3feat} adopt the joint descriptor and detector paradigm from D2-Net and extended it to point clouds. 
Their descriptor 
is trained using a contrastive loss, while a self-supervised loss is proposed for the detector, based on the descriptor's response. 
In contrast to these works we only train the keypoint detector while using a pre-existing descriptor.

\subsection{Saliency Estimation}

\emph{Saliency} has been studied as a way to quantify and understand what is relevant within a neural network. The definition of 
saliency in this context  has been studied by a number of 
works, such as ~\cite{Simonyan14a,Selvaraju_2017_ICCV,NIPS2018_8160} as a way to explain why machine learning models 
behave as they do. Even before the growth in popularity of neural networks, 
methodologies such as~\cite{BaehrensJMLR10} were being designed to understand why classifiers made specific decisions, but 
became more prevalent with the adoption of deep learning models for most perception tasks. Approaches such 
as~\cite{Simonyan14a,NIPS2018_8160} studied what a neural network finds relevant by looking at how the gradients 
of a given prediction behave. Their findings showed that gradients with higher magnitude in specific areas influence predictions.

The use of saliency has been demonstrated to be capable of generating state-of-the-art keypoint extractors in images. 
ELF~\cite{elf_ICCV19} computed the gradient of the feature map given an image and then used the Kapur threshold~\cite{Kapur1985} 
to select keypoints. In a similar fashion, Grad-CAM~\cite{Selvaraju_2017_ICCV} used the gradient maps of a classification score to 
produce regions of interest for a given image for tasks such as classification, image captioning, and visual question 
answering. Inspired by 
these works, our approach applies saliency ideas to extract reliable keypoints 
on point cloud data.

\begin{figure*}[t!]
	\begin{center}
		\vspace{2mm}
		\includegraphics[width=0.95\textwidth]{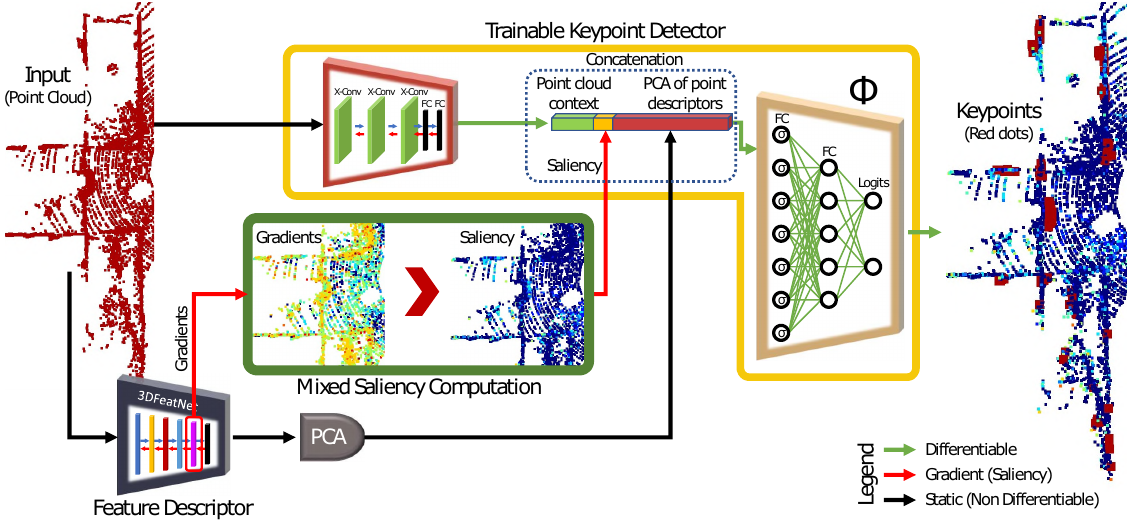}
	\end{center}
	\vspace{-4mm}
	\caption{Proposed network architecture. The input to our system is a raw point cloud and a feature 
		descriptor 
		network. Our method concatenates the \emph{saliency} information from the gradients of the 
		feature descriptor 
		with the per-point context features and a PCA projection of the original feature descriptor. The 
		concatenated 
		vector is the input to two FC layers that generate per-point response at the logits layer.}
	\label{fig:architecture}
	\vspace{-6mm}
\end{figure*}

\section{Methodology}
\label{sec:method}

We present Salient Keypoint Detection (\textbf{SKD}) - a keypoint extraction method based on the \emph{saliency} of a point 
cloud. Our approach follows \emph{describe-then-detect} methodology to extract keypoints based on a pre-existing 
differentiable descriptor by the means of \emph{saliency}. Both are combined with context information to using a neural network 
that learns to predict the likelihood of a point being a keypoint (\secref{sec:neural_network}). \figref{fig:approach_qualitative} 
shows the main steps of our detector, which are further explained in the following sections.

\subsection{Point Cloud Saliency}
\label{sec:saliency}

Given a point cloud $P = \{ p_i \in \mathbb{R}^3 \}^N$ with $N$ points and a pre-trained descriptor network $F$, we extract the 
gradients of the network at a specific layer $l$ with respect to the positional information of the input point cloud, defined as 
$\nabla F^P_l \in \mathbb{R}^{\Omega_l \text{x} N \text{x} 3}$, where $\Omega_l$ is the feature map domain at layer 
$l$. 

We define \emph{initial saliency} $S(P)$, as the product of the feature activations of that layer  $F^P_l \in 
\mathbb{R}^{\Omega_l}$ with the gradients,  $\nabla F^P_l$. This is formally defined as:

\begin{equation}
\label{eq:saliency}
S(P) = F^P_l \cdot \nabla F^P_l
\end{equation}

where, $S(P) \in \mathbb{R}^{N \text{x} 3}$.
In this way the extracted initial saliency corresponds to specific points in the point cloud that have good 
activations and are valuable based on the gradient of that layer w.r.t. the input point cloud. From a geometric perspective this 
can be thought of as projecting the feature signal through to the input point cloud and determining the value of 
individual points (see example projection in~\figref{fig:approach_qualitative}).

We take the initial saliency computed from~\eqref{eq:saliency} and determine a secondary saliency score per point, defined as 
$s_i$ for each point. The score weighs the contribution of $S(P)$ by the distance from the median of the point 
cloud~\cite{pointCloudSal_ICCV19}. The score is formally defined as:

\begin{equation}
	s_i = -\sum_{j=1}^{3}\big[S_j(P) \cdot (x_{ij} -\mathit{median}(x_{ij}))\big]r_i
\end{equation}

where $j \in \{1,2,3\}$ defines each of the Cartesian coordinates $x,y,z$ of point $p_i \in P$ and 
$r_i=\sqrt{\sum_{j=1}^{3}(x_{ij}-\mathit{median}(x_{ij}))^2}$  is the distance 
of point $i$ to the median of the point cloud. Therefore, further away points will have higher score, and points closer to the 
center will have lower 
score (see~\figref{fig:approach_qualitative}). This design decision is motivated by the high density of points near the origin of the sensor (see 
distribution of points of 3DFeatNet vs SKD in~\figref{fig:kitti_qualitative}). Also, note that well 
distributed keypoints in two point clouds will likely result in a more robust solution.

Finally, in accordance with standard deep learning practices, the input saliences are normalized to have zero mean and unit variance within a 
single point cloud. The final saliency ensures a good spatial distribution of the selected points while also selecting points with good activations 
based on the descriptor network.~\figref{fig:approach_qualitative} illustrates the saliences of an example input point cloud at each stage of 
computation.


\subsection{Network Architecture}
\label{sec:neural_network}

In order to determine the keypoints, we designed a neural network to combine 
the saliences with other sources of information. Our architecture is depicted 
in~\figref{fig:architecture} and consists of three modules that generate a 
combined feature vector, which is the input for a final fully-connected network 
used to compute the likelihood of each point being a keypoint.

\begin{itemize}
	\item The first component, \emph{mixed saliency computation}, is the 
	\textit{saliency} method described in the previous section. 
	\item The second component is a \emph{PCA dimensionality reduction} of the 
	original per-point features. This improved the final results by producing a 
	smoother feature space.
	\item The third part is called the \emph{point cloud context} features. 
	These are four X-Conv layers~\cite{pointCNN_NIPS18} and two 
	fully connected (FC) layers that have been pretrained on a feature 
	extraction task to create stable initial estimates. 
	These layers learn to provide a description of the context around any 
	given point. We use a two dimensional size for the context latent space. 
\end{itemize}

The output of the three components are concatenated and fed to two additional 
fully connected layers, denoted by $\Phi$, to produce the final keypoint 
prediction. The network learns to infer a score for each point --- determining 
the likelihood of it 
being a robust and repeatable keypoint for the 
original descriptor. Note that our model is descriptor-agnostic and thus can be applied to any descriptor network in 
order to improve the performance.

\subsubsection{Training}

During training, the input to our model consists of a pair of point clouds and the ground truth 
transformation between them, $(P_k,P_l,T)$. Both $P_k$ and $P_l$ are $N \times 3$ dimensional 
vectors, where $N$ denotes the cardinality of the point cloud set. In addition, we assume to have a 
pre-trained model for the point descriptors. To train we estimate the saliency $s_i$ and features $f_i$ for each point in both 
point clouds. Due to the large dimensionality of the feature space $f_i$, we performed PCA and 
transformed the features that explained $\approx 90\%$ of the data.
In addition, given the ground truth transformation between the two point clouds, we determine the bidirectional correspondence for each point, 
conditioned on the descriptor. These 
correspondences are used in our loss to select matching pairs of points. These 
correspondences need not be injective nor surjective: if for a point $i_1$ the closest 
neighbour under the ground truth transformation (up to a \SI{1}{\meter} threshold) is $j_1$, it does not mean the reverse 
applies nor that $i_1$ has a unique neighbour from $P_l$.

For each point in both point clouds we extract  \textit{context-aware features}, $f_c$, 
and concatenate them to the saliencies $s_i$ and features $f_i$. The term \emph{context-aware} is used 
as we expect the layers that lead to this descriptor to contain information about the local geometry
around each point. We pretrain these layers on a feature learning task to obtain stable initial features in the training. We chose a small feature 
space of only two dimensions in order to force the network to learn rough estimates of the shape of objects that can generalize better when 
moving to a different dataset. Afterwards, the full architecture is trained end-to-end with the saliency and feature concatenation. We note that 
the contribution of the \textit{context-aware features} is smaller compared to $s_i$ and $f_i$.

The concatenated saliences, PCA features, and \emph{context-aware} features are fed to two fully 
connected layers to estimate the likelihood of each of the points being a keypoint. To this end we use a softmax 
cross entropy loss between the stacked $P_k, P_l$ clouds and the determined correspondences, given the ground truth 
transformation. Due to the smaller number of correct keypoint correspondences between the two clouds, we balance 
the loss function terms given the keypoint to non-keypoint ratio determined by the ground truth correspondences. 

\subsubsection{Inference}

During the forward pass of the network we estimate a likelihood for each point being a correct keypoint. 
We can then either extract the top $K$ keypoints based on this likelihood or select all the keypoints 
above certain threshold. The extracted keypoints produced by the described approach are 
better suited to the descriptor as the learning iterations optimize its performance. Note, that neither Non-Maximum 
Suppression, nor any other threshold is applied to filter the final set of keypoints.


\section{Results}
\label{sec:results}

Next, we discuss the datasets, baselines, and metrics we used to evaluate our approach and then present our findings.

\begin{figure}[t]
	\begin{center}
		\vspace{2mm}
		\includegraphics[trim={0 1mm 0 0},clip,width=0.9\linewidth]{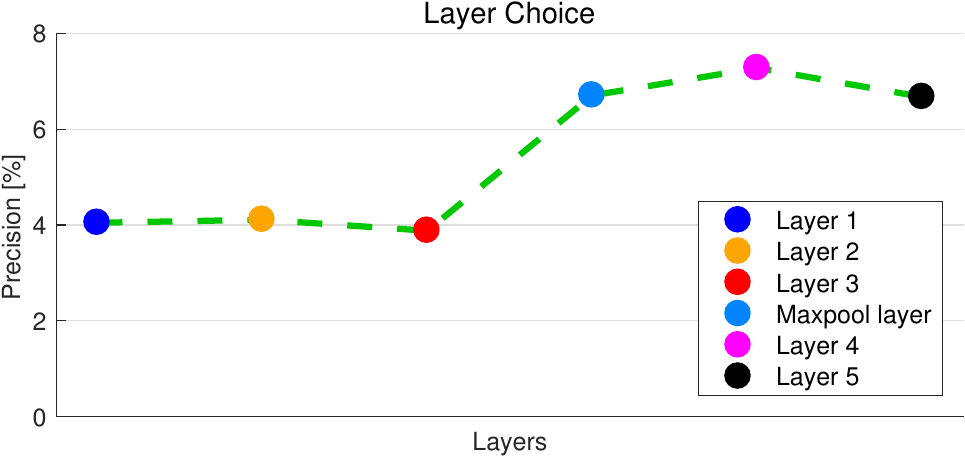}
	\end{center}
	\vspace{-4mm}
	\caption{Performance of each layer of 3DFeatNet expressed as percentage of correct matches if the layer was selected for 
		gradient computation. The layer identified corresponds to the counter of layers in~\cite{3dfeatnet} starting from the left. In 
		our evaluation we used the layer that maximized the performance (Layer 4) for both ELF~\cite{elf_ICCV19} and our method.}
	\label{fig:layer_choice}
	\vspace{-7mm}
\end{figure}

\subsection{Datasets}

In our study we used two datasets - the Oxford RobotCar~\cite{RobotCarDatasetIJRR} 
and the KITTI odometry dataset~\cite{kitti2013IJRR}. In order to provide a fair comparison, our 
experiments use the preprocessed test data provided by related previous works, ~\cite{3dfeatnet} 
and~\cite{li2019usip}. We also used their evaluation scripts to make the comparison as clear as 
possible. We trained our model using the same sequences from the RobotCar dataset 
as~\cite{3dfeatnet}, and tested our approach using the same test set of $3,426$ point cloud pairs which 
the authors provided. Furthermore, we neither trained our method nor these baselines on the KITTI dataset, 
in order to test the generalization ability of the approaches. 

The evaluation part of the KITTI dataset, used by both~\cite{3dfeatnet,li2019usip}, provides only 
$2,369$ point clouds out of the total dataset. So as to increase the size of the KITTI evaluation 
dataset, we extended it using the 11~\emph{training} sequences. This is possible only because the 
RobotCar dataset is used for training all models. The extended dataset is processed in a dense 
manner: for each point cloud we align the next consecutive 10 point clouds to it using the ground truth 
sensor poses. By doing this we expanded the number of testing point cloud pairs from $2,831$ to 
$207,917$, which allows us to more fully study our proposed approach. Note that none of the KITTI data was used for 
training our method, any of the baselines, or the descriptor.

\begin{figure*}[t]
	\begin{center}
		\vspace{2mm}
		\includegraphics[width=0.32\textwidth]{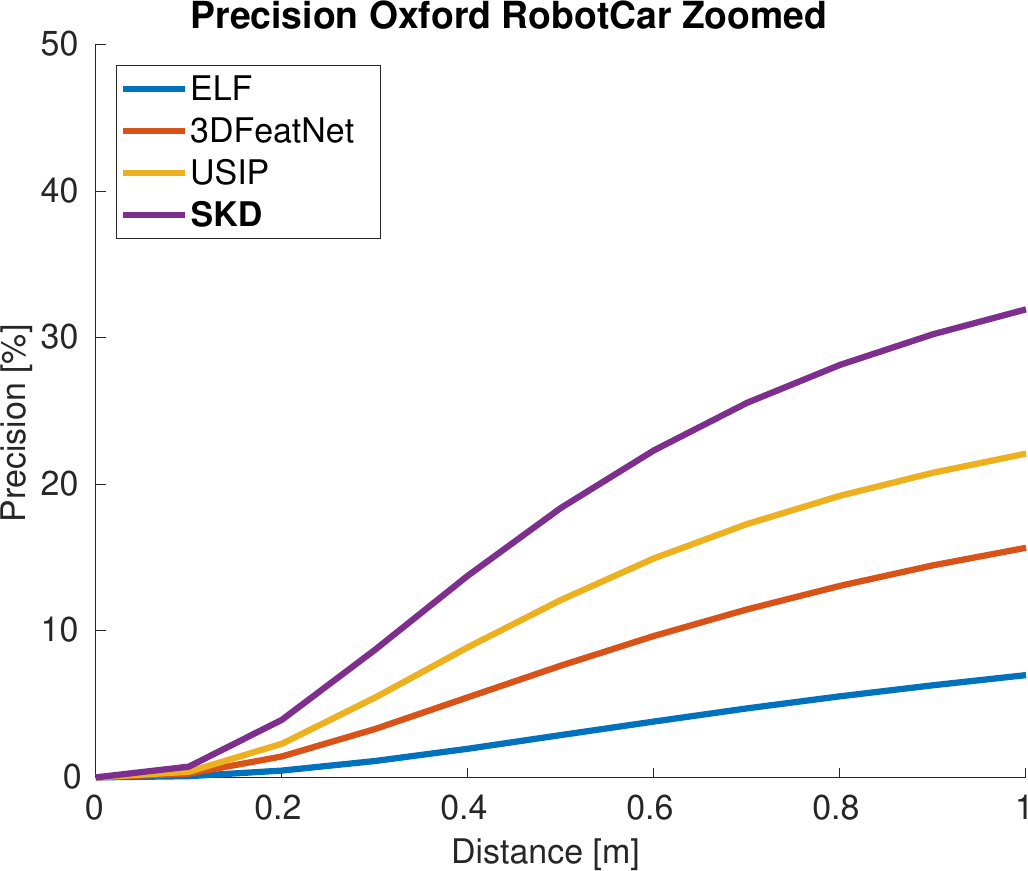}
		\includegraphics[width=0.32\textwidth]{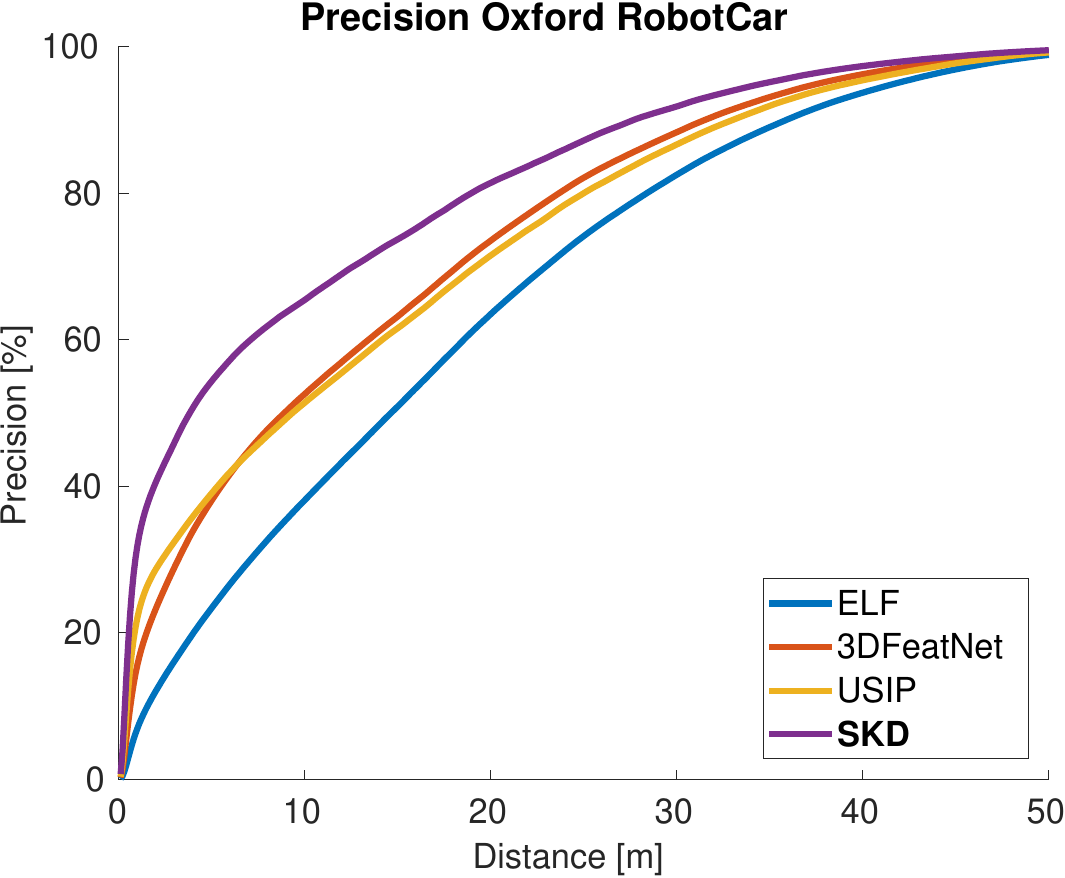}
		\includegraphics[width=0.32\textwidth]{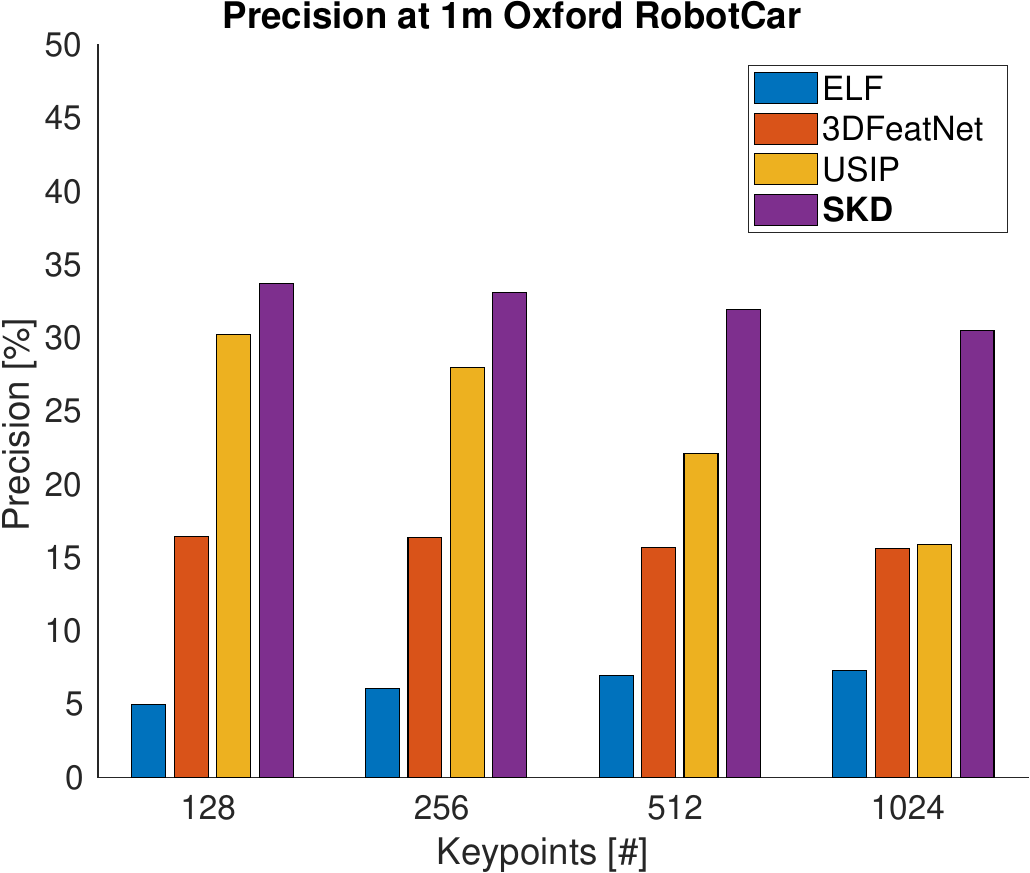}
	\end{center}
	\begin{center}
		\includegraphics[width=0.32\textwidth]{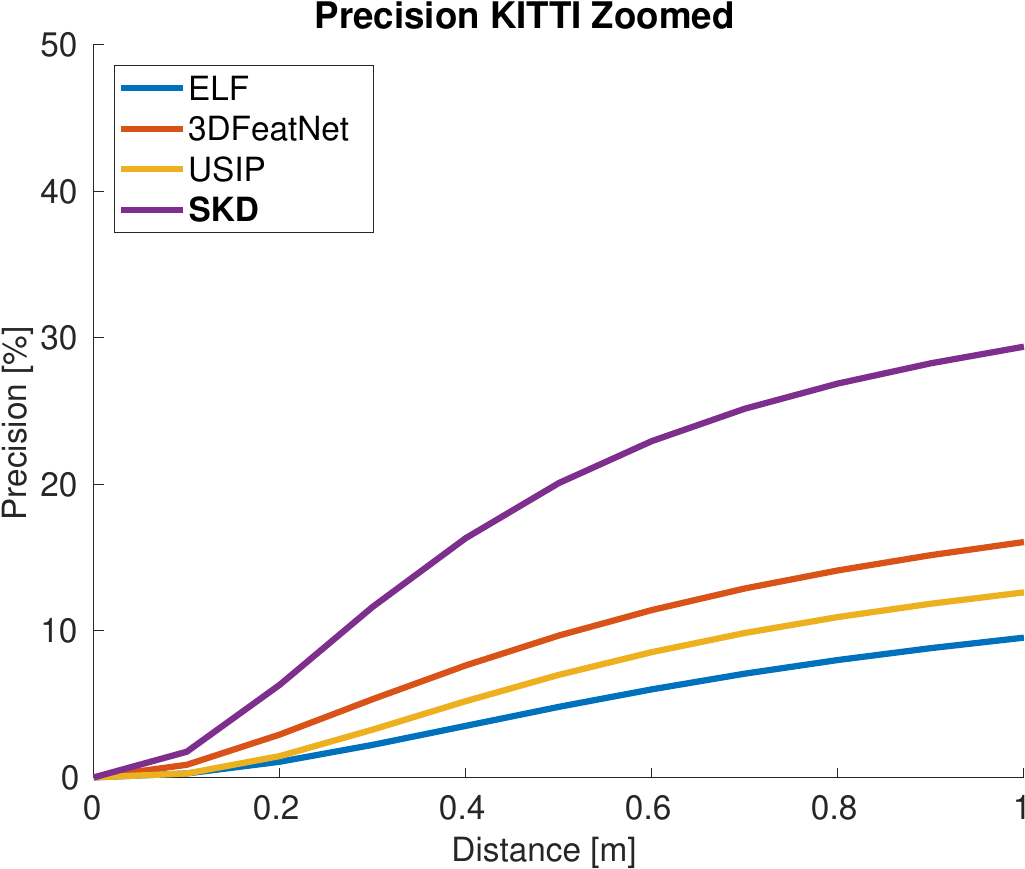}
		\includegraphics[width=0.32\textwidth]{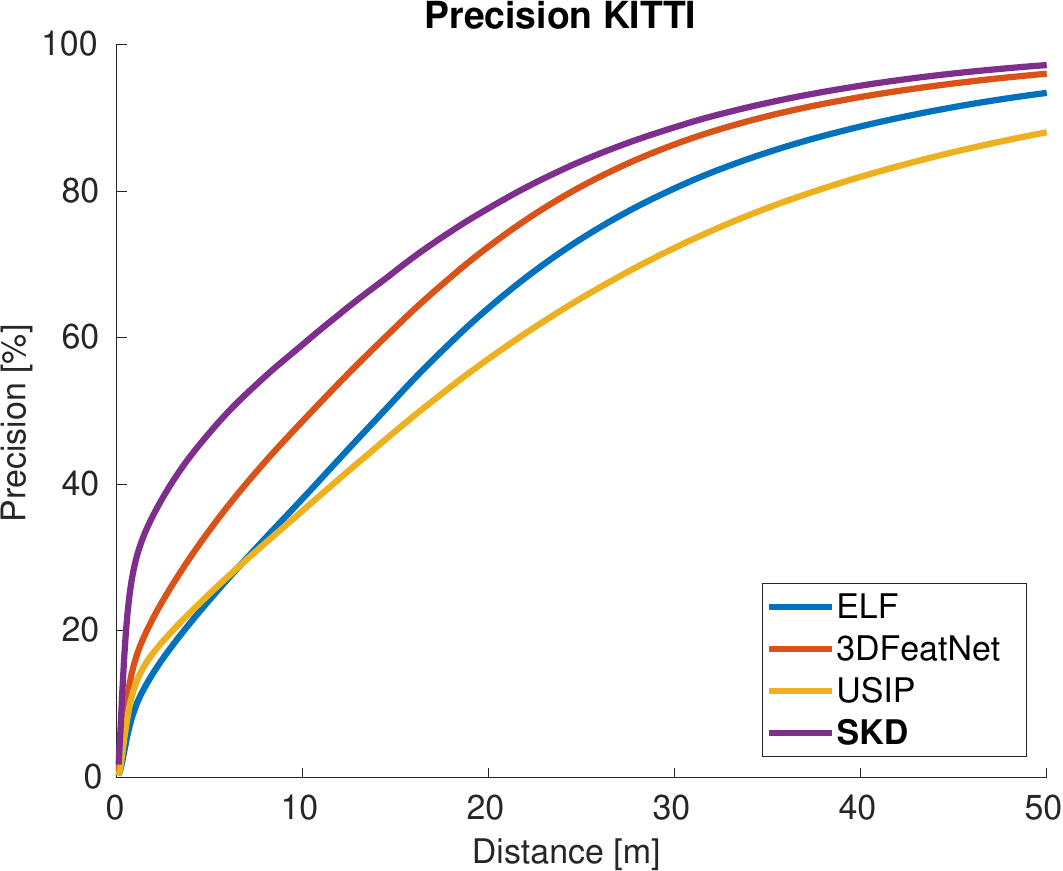}
		\includegraphics[width=0.32\textwidth]{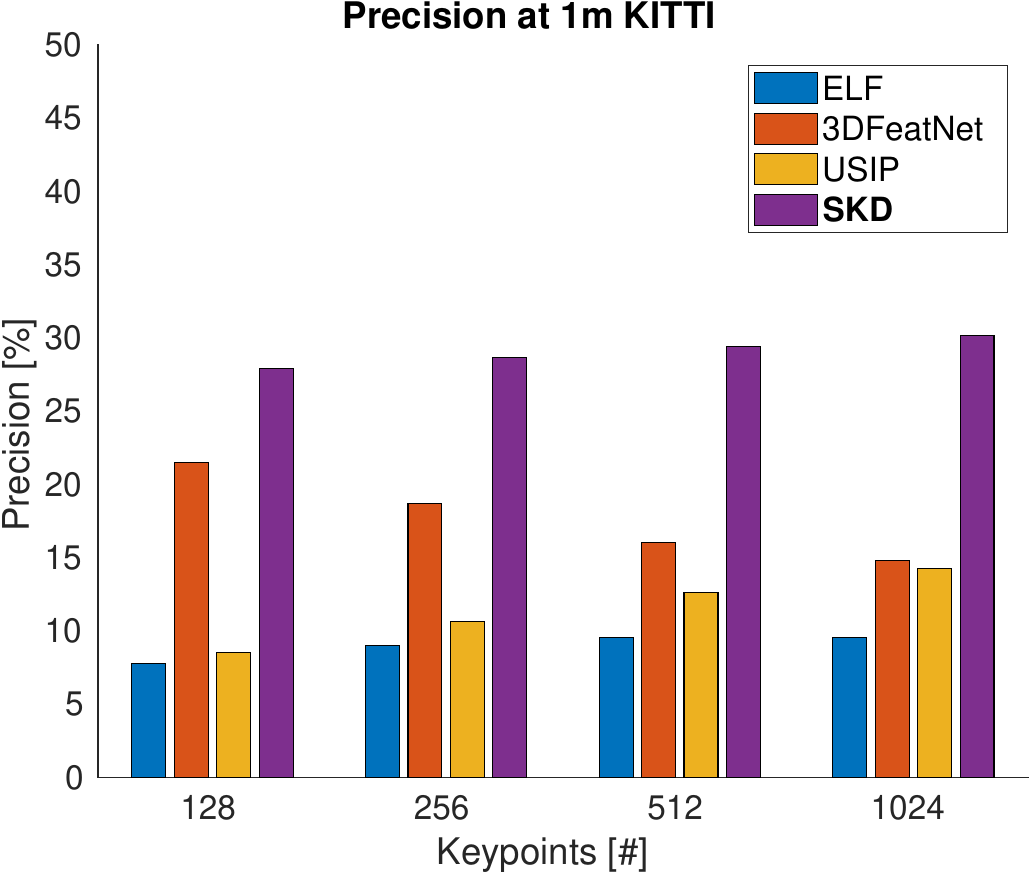}
	\end{center}
	\caption{Matching score evaluated on the Oxford RobotCar dataset (top row) and the KITTI dataset 
		(bottom row). 
		The first two columns are the percentage of matched keypoints using the 
		3DFeatNet~\cite{3dfeatnet} 
		descriptor when 
		varying the distance between correspondences. The first column shows a zoomed-in version of the 
		second column 
		within \SI{1}{\meter} of distance, considered relevant for geometric registration. The third column 
		shows the 
		performance of different approaches while varying the number of keypoints.}
	\label{fig:precision}
	\vspace{-6mm}
\end{figure*}

\subsection{Metrics}

We focused our analyses on the keypoint extraction methods and compared them using three metrics. To ensure a fair 
comparison to the baseline keypoint extraction methods (see~\secref{sec:layer_justification}), we used the same 
3DFeatNet~\cite{3dfeatnet} descriptor for all the methods.

The first metric used the \emph{matching score} proposed by~\cite{3dfeatnet} that determined correspondences within 
\SI{1}{\meter} maximum distance. We detected 
keypoints separately for two point clouds. Given the ground truth sensor poses, we projected the 
keypoints from the first point cloud into the second one. Keypoints that did not have a nearest 
neighbour bidirectional correspondence in the second point cloud were ignored from the final result (i.e. no overlap). For the 
rest of 
the keypoints, the descriptors were compared and matched to establish a correspondence. The 
precision was measured as the number of correct correspondences against the total number of possible 
matches. This metric estimated the percentage of correct correspondences based on the distance 
between them. Correct correspondences were thresholded to \SI{1}{\meter} distance, as it 
served as a desirable upper limit when performing registration between two point clouds.

For the second metric we chose to compare the \emph{normalized relative repeatability}, as proposed 
in~\cite{li2019usip}. The metric compared the keypoints detected in one point cloud to the keypoints detected in its 
corresponding point cloud. If the keypoints overlapped, below a certain distance, they were considered a 
match. 
While this saturated with a high number of keypoints~\cite{elf_ICCV19}, we chose to use this metric to have a fair 
comparison against USIP~\cite{li2019usip}. We note that the ability to match keypoints on their own is 
not sufficient, hence we considered the matching score to be the more important metric.

Finally, we evaluated the \emph{geometric registration performance} of our approach on both the KITTI and the Oxford 
RobotCar datasets using RANSAC~\cite{ransac}. As in~\cite{3dfeatnet}, we consider an alignment to be successful 
if it differs by less than \SI{2}{\meter} and \SI{5}{\degree} from the ground 
truth 
sensor pose. For the 
successful registrations we present relative translation error (RTE), relative rotation error (RRE), 
success rate as a percentage of the successful registrations, average 
number of iterations it took RANSAC to find a suitable candidate within 99\% confidence (capped at 
$10,000$ iterations), and the inlier ratio of how many points were considered when obtaining a 
correct registration. 

\subsection{Baseline Methods}
\label{sec:layer_justification}

Our approach is general and can be applied to any point cloud descriptor network. For simplicity we chose to use the 
descriptor~\cite{3dfeatnet} as it is open source and easy to use, as well as being state-of-the-art point cloud descriptor. We 
select the best performing layer from the descriptor to generate the gradients on which we compute 
the saliency values, as illustrated in~\figref{fig:layer_choice}.
We compared against the learned keypoint detector methods of 3DFeatNet~\cite{3dfeatnet}, 
DH3D~\cite{dh3d}, USIP~\cite{li2019usip}, and a 3D interpretation 
of the ELF~\cite{elf_ICCV19} detector. All the learning methods were trained on the Oxford RobotCar 
dataset and tested on both RobotCar and KITTI data. We have used the models provided online, and 
trained our own network. For USIP, we took the models provided by the authors which they trained on the Oxford 
RobotCar dataset. As ELF does not need training, we took the best performing layer, 
according to~\figref{fig:layer_choice}. We adapted their approach to work on point clouds by 
performing Non-Maximum Suppression in 3D and choosing keypoints based on the Kapur 
Threshold~\cite{Kapur1985}.

\subsection{Matching Score Experiments}

We evaluated our approach using the matching score metric on both Oxford RobotCar and KITTI 
datasets.~\figref{fig:precision} illustrates the performance of SKD compared to other state-of-the-art 
methods. The first row shows the percentage of matched keypoints on the Oxford RobotCar dataset 
using the 3DFeatNet descriptor for all methods. The first column is a zoomed-in version of the 
percentage of points detected within \SI{1}{\meter} distance that are considered relevant for matching. 
In the second column we show the full results as presented in~\cite{3dfeatnet}.
The last column shows the precision at \SI{1}{\meter} distance at different numbers of keypoints --- 
128, 256, 512, 1024. As all the approaches are learning-based, we took the top $K$ keypoints; this 
ensures that the number of keypoints are identical. The second row presents the results for the full KITTI 
dataset, with all the methods trained on the Oxford RobotCar dataset. We observe that our approach 
generalizes well without a loss in performance, and outperforms the second best approach by a 
significant margin. Similarly to 3DFeatNet~\cite{3dfeatnet}, our approach's performance does not 
decline with the increase of detected keypoints.

\begin{figure}[t]
	\begin{center}
		\vspace{2mm}
		\includegraphics[trim={0 0.1cm 0cm 0cm},clip,width=0.49\linewidth]{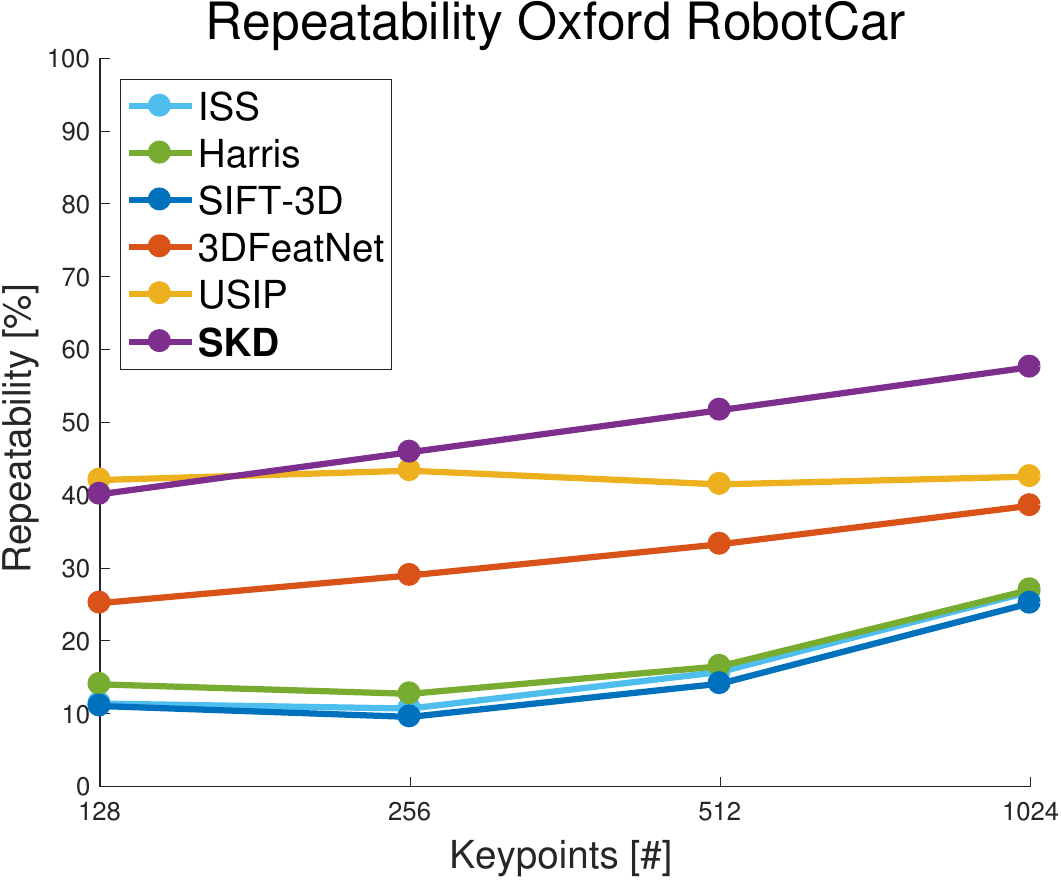}
		\includegraphics[trim={0 0.1cm 0cm 0cm},clip,width=0.49\linewidth]{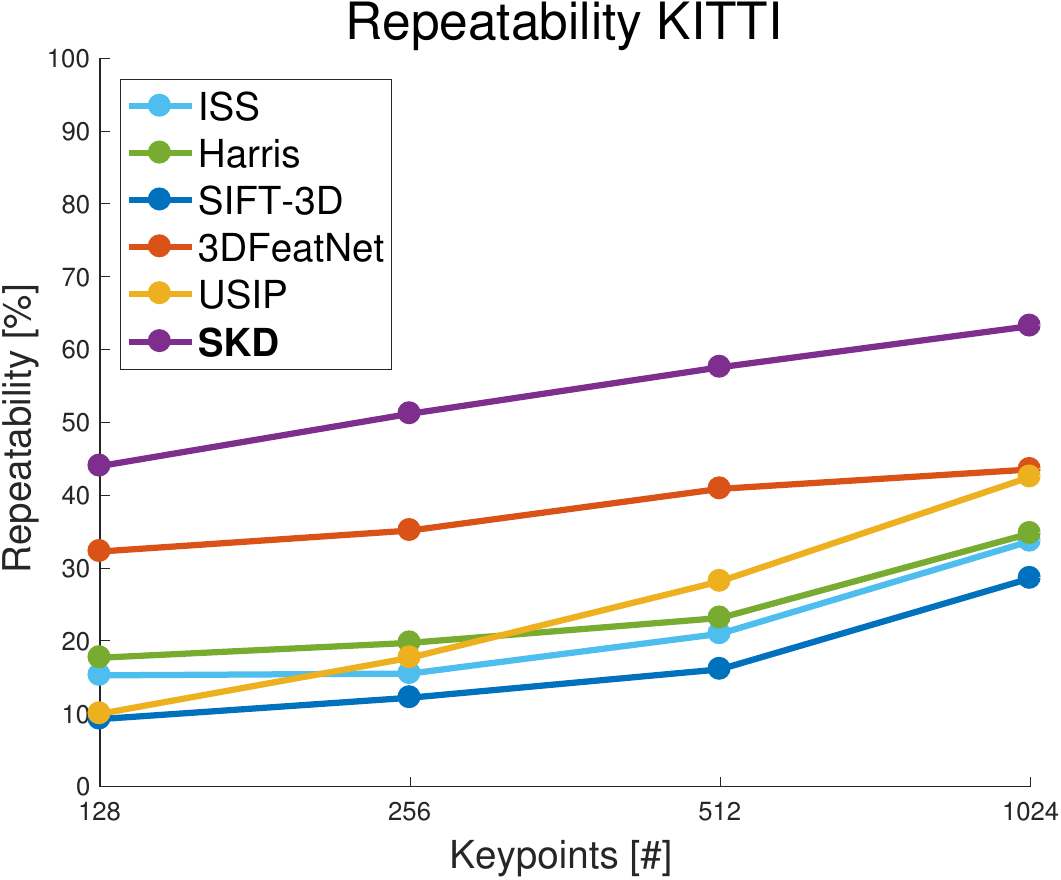}
	\end{center}
	\vspace{-6mm}
	\caption{Relative repeatability on the Oxford RobotCar dataset (left) and 
		KITTI (right) with different number of extracted keypoints. }
	\label{fig:repeatability}
	\vspace{-6mm}
\end{figure}

\subsection{Repeatability Experiments}

\begin{figure}[t]
	\begin{center}
		\includegraphics[trim={0 13mm 0 0},clip,width=1.0\linewidth]{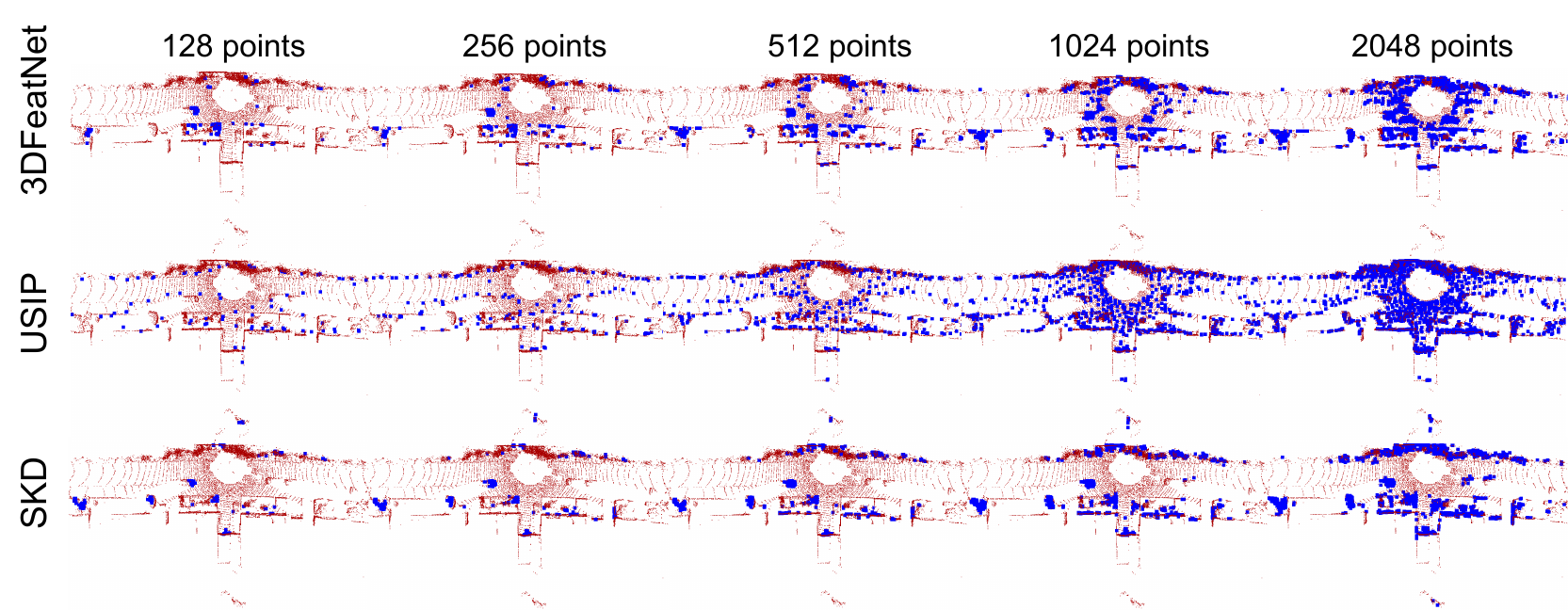}
	\end{center}
	\vspace{-4mm}
	\caption{Qualitative results on the KITTI dataset. For all methods the number of generated keypoints is 
		increased from left to right. This shows what relative importance each method gives to certain areas 
		of the point cloud.}
	\label{fig:kitti_qualitative}
	\vspace{-6mm}
\end{figure}

\begin{figure}[t]
	\begin{center}
		\vspace{2mm}
		\includegraphics[trim={0 3mm 0 0.2cm},clip,width=1.0\linewidth]{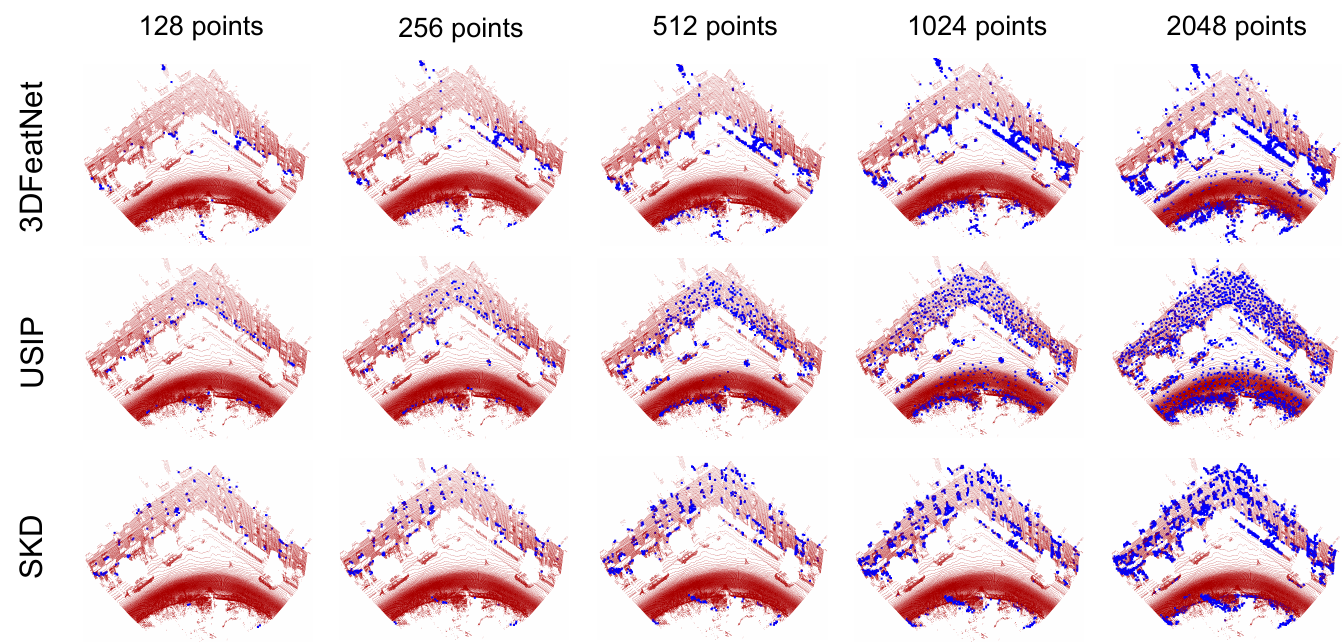}
	\end{center}
	\vspace{-4mm}
	\caption{Qualitative results on the Oxford RobotCar dataset. The keypoint proposals from SKD are evenly spread in 
		geometrically interesting locations.}
	\label{fig:oxford_qualitative}
	\vspace{-6mm}
\end{figure}

Next, we evaluate the repeatability of keypoints across multiple observations of the same scene. ~\figref{fig:repeatability} shows 
the repeatability of SKD measured on the Oxford RobotCar 
dataset (left) and the KITTI dataset (right). For the Oxford RobotCar, our approach performs 
comparably to the second best method when extracting 128 and 256 keypoints, but it selects points that 
are 40\% more repeatable when extracting 1024 interesting points. 
Furthermore,~\figref{fig:repeatability_visual},~\ref{fig:kitti_qualitative},~\ref{fig:oxford_qualitative}, and the supplementary video 
material present qualitative 
examples of the detection of keypoints of the top three approaches. The proposed approach has learned 
to select more descriptive areas of the environment. For example, our method 
does not select ground points, and even though we forced the method to generate large amounts of 
points, it still chooses points based on the high activations of the network - around building edges and 
corners, that in turn are easier to match.

~\figref{fig:repeatability} (right) illustrates the repeatability of our approach on the KITTI dataset. On 
this dataset, 3DFeatNet performs slightly better than on the Oxford RobotCar dataset. Interestingly, as 
we use the feature signal of 3DFeatNet, our keypoint detector also performs better. In 
addition,~\figref{fig:kitti_qualitative} presents qualitative 
evaluation of our approach on the KITTI dataset with an increased number of keypoints, where SKD 
consistently selects the same areas of the environment.

\subsection{Geometric Verification Experiments}

Although our work is focused on keypoint detection, we show results for one possible application of our detector --- 
geometric registration.~\tabref{tab:geom_errors} presents the results of geometric registration on KITTI and Oxford RobotCar. 
We used the baselines' official code and models, but note that we used 73 times more evaluation data for KITTI.

SKD performs similarly to the state-of-the-art in terms of relative rotation and translation error, within the standard 
deviation of the best-performing method (\SI{2}{\milli\meter} and \SI{0.04}{\degree} difference with best baseline). Our method is, 
however, 
approximately nine times faster on the KITTI dataset and two times faster on the Oxford RobotCar dataset compared to the 
second-best approach with a higher percentage of inliers. We note that we did not use non-maximum 
suppression (NMS), which we suspect is the reason for the slightly higher errors and lower success rate. Applying NMS first 
and then extracting the top K keypoints would reduce the number of grouped keypoints, and as such the robust estimator 
could find better distributed keypoints, which will likely result in better point cloud registration accuracy. As shown in our 
\textit{repeatability} and \textit{matchability} experiments, our method is good at picking out keypoints at similar locations. 
However, for applications such as global registration highly repeatable and matchable points are not the only necessary 
requirement. In contrast, for object pose estimation highly repeatable points are arguably more important than the 
distribution of points. Furthermore, not using NMS makes our detector more robust to changes in the scene.

\begin{table*}[t]
	\vspace{2mm}
	\resizebox{\textwidth}{!}{
		\begin{tabular}{|c|c||cccccc|}
			\hline
			& Detector + Descriptor Method
			&  & RTE (m) & RRE ($^{\circ}$) & Success Rate & Avg \# iter & Inlier ratio  \\ \hline
			\parbox[t]{2mm}{\multirow{5}{*}{\rotatebox[origin=c]{90}{Oxford}}} 
			& ELF~\cite{elf_ICCV19} + 3DFeatNet~\cite{3dfeatnet} $\ast$ &  & $0.42 \pm 0.31$ & $1.66 \pm 1.09$ 
			& 
			$86.49\%$ & $9788$ & $5.3\%$ \\
			& 3DFeatNet~\cite{3dfeatnet} + 3DFeatNet~\cite{3dfeatnet} $\ast$ &  & $0.30 \pm 0.25$ & $1.07 \pm 
			0.85$ & $97.64\%$ & $3083$ & 
			$12.9\%$ \\
			& USIP~\cite{li2019usip} + 3DFeatNet~\cite{3dfeatnet} $\ast$ &  & $0.29 \pm 0.26$ 
			& 
			$\mathbf{0.96} \pm \mathbf{0.77}$ 
			& 
			$\mathbf{98.74\%}$ 
			& 
			$823$ & 
			$21.0\%$\\ 
			& DH3D~\cite{dh3d} + 3DFeatNet~\cite{3dfeatnet} &  & $\mathbf{0.28}$ 
			& 
			$1.04$ 
			& 
			$98.2\%$ 
			& 
			$2908$ & 
			-\\ 
			\cline{2-8}
			& \textbf{SKD} + 3DFeatNet~\cite{3dfeatnet} &  & $0.31 \pm 0.27$ & $1.11 \pm 0.89$ & 
			$97.64\%$ 
			& 
			$\mathbf{393}$ & 
			$\mathbf{32.7\%}$\\ 
			\hline
			\hline			
			\parbox[t]{2mm}{\multirow{3}{*}{\rotatebox[origin=c]{90}{KITTI}}} & 3DFeatNet~\cite{3dfeatnet} + 
			3DFeatNet~\cite{3dfeatnet}  $\ast$ &  & $0.142 \pm 0.120$ & $\mathbf{0.533} \pm 
			\mathbf{0.410}$ & $\mathbf{97.80\%}$ & $3917$ & $12.7\%$ \\
			
			& USIP~\cite{li2019usip} + 3DFeatNet~\cite{3dfeatnet} $\ast$&  & $0.203 \pm 0.193$ 
			& $0.637 \pm 0.517$ & $97.12\%$ & $5324$ & $11.0\%$\\ 
			\cline{2-8}
			
			& \textbf{SKD} + 3DFeatNet~\cite{3dfeatnet} &  & $\mathbf{0.140} \pm \mathbf{0.134}$ & $0.579 \pm 0.480$ & 
			$96.52\%$ &	$\mathbf{594}$ & $\mathbf{32.2\%}$\\ 
			\hline
		\end{tabular}
	}
	\caption{Geometric registration on the Oxford RobotCar dataset (top) and the KITTI dataset (bottom), evaluated 
		by RANSAC. $\ast$ indicates our evaluation on the dataset using each method's official implementation 
		and models. Note: the used KITTI dataset is 73 times larger than the one by~\cite{3dfeatnet,li2019usip}.}
	\label{tab:geom_errors}
	\vspace{-10mm}
\end{table*}

\subsection{Ablation studies}

\begin{figure}[t]
	\begin{center}
		\includegraphics[trim={0 0 0 0.4cm},clip,width=1.0\linewidth]{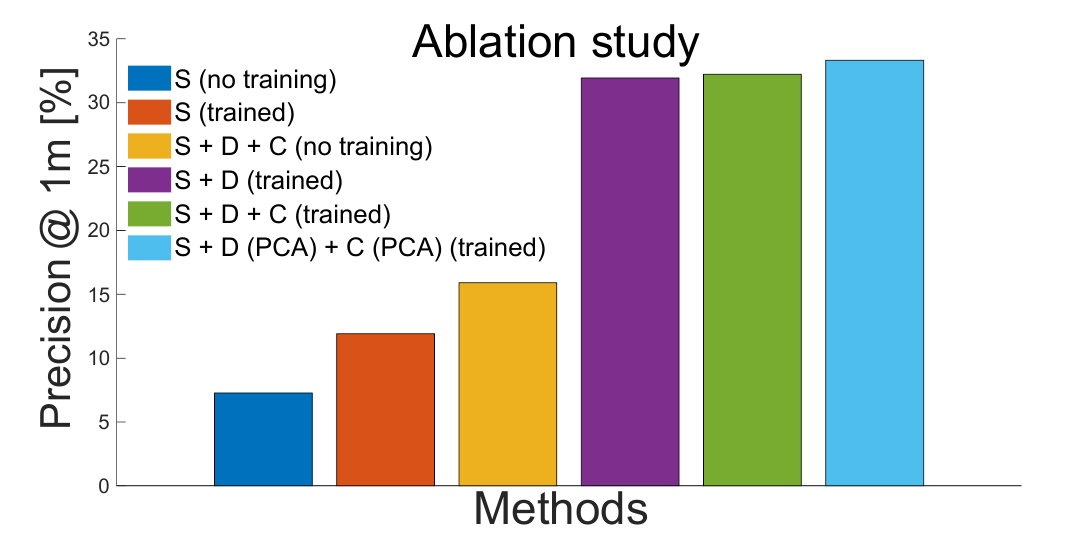}
	\end{center}
	\vspace{-5mm}
	\caption{S=Saliency, D=Descriptor, C=Context. Ablation studies of our method evaluated on the Oxford RobotCar dataset. We 
	evaluate the 
		contribution of the saliency information, and the importance of training with the original descriptor. 
		The results displayed in the figure show the clear contribution the neural network is making to the 
		keypoint generation.}
	\label{fig:ablation}
	\vspace{-7mm}
\end{figure}

Next, we studied the contribution of each component of our method by measuring the precision at \SI{1}{\meter} distance on 
the Oxford RobotCar dataset, shown in~\figref{fig:ablation}. Note that the saliency can be extracted without any training and 
thus we present ablation studies with both options.

First, we inspected the contribution of the saliency in the concatenated vector. In our 
experiments the saliency information alone is responsible for approximately half of the concatenated 
vector - dark blue compared to yellow in~\figref{fig:ablation}. In addition, training the neural network, $\Phi$, with 
just saliency information contributes to approximately a third of the end result - red 
compared to light blue, thus justifying the name of our method. Second, the contribution of the network, $\Phi$,  when training 
with the full feature 
vector, is substantial as is to be expected - purple compared to dark blue and yellow 
on~\figref{fig:ablation}. By analyzing our ablation study in~\figref{fig:ablation} we can see that
the increase in performance from the context features is only marginal. We would like to note that the performance gains, 
and thus the contributions of this paper, come from using saliency together with the 
value of the feature to determine which points to extract as keypoints. Finally the use of PCA was introduced 
to smooth the feature space and allow for better generalization, and improved performance, as can be seen in 
in~\figref{fig:ablation} (green vs light blue).

\section{Conclusions}
\label{sec:conclusions}

In this paper we presented a novel method for keypoint extraction that uses saliency information to extract informative regions in 
a point cloud. The method concatenates saliency computed from the gradient of a pre-trained descriptor network evaluated at 
the input cloud, context-aware features, and the descriptor features and learns to predict which 3D points have a higher chance 
of being matched correctly. The proposed approach is descriptor-agnostic and outperforms the state-of-the-art by up to 50\% 
in matchability and repeatability compared to the second-best method. When performing geometric registration, the algorithm 
achieves a higher number of inliers while also finding a correct pose estimate significantly faster than the current 
state-of-the-art.

\section{Acknowledgments}
\label{sec:acknowledgments}

We would like to thank Tom Dobra for maintaining the resources used for this project. We are also grateful for the detailed 
corrections by Matias Mattamala.

\small
\bibliographystyle{IEEEtran}
\bibliography{egbib}

\begin{thebibliography}{10}
\providecommand{\url}[1]{#1}
\csname url@samestyle\endcsname
\providecommand{\newblock}{\relax}
\providecommand{\bibinfo}[2]{#2}
\providecommand{\BIBentrySTDinterwordspacing}{\spaceskip=0pt\relax}
\providecommand{\BIBentryALTinterwordstretchfactor}{4}
\providecommand{\BIBentryALTinterwordspacing}{\spaceskip=\fontdimen2\font plus
\BIBentryALTinterwordstretchfactor\fontdimen3\font minus
  \fontdimen4\font\relax}
\providecommand{\BIBforeignlanguage}[2]{{%
\expandafter\ifx\csname l@#1\endcsname\relax
\typeout{** WARNING: IEEEtran.bst: No hyphenation pattern has been}%
\typeout{** loaded for the language `#1'. Using the pattern for}%
\typeout{** the default language instead.}%
\else
\language=\csname l@#1\endcsname
\fi
#2}}
\providecommand{\BIBdecl}{\relax}
\BIBdecl

\bibitem{dusmanu2019d2}
M.~Dusmanu, I.~Rocco, T.~Pajdla, M.~Pollefeys, J.~Sivic, A.~Torii, and
  T.~Sattler, ``{D2-Net:} a trainable {CNN} for joint description and detection
  of local features,'' in \emph{CVPR}, 2019, pp. 8092--8101.

\bibitem{r2d2}
J.~Revaud, P.~Weinzaepfel, C.~R. de~Souza, and M.~Humenberger, ``{R2D2:}
  repeatable and reliable detector and descriptor,'' in \emph{NIPS}, 2019.

\bibitem{elf_ICCV19}
A.~Benbihi, M.~Geist, and C.~Pradalier, ``{ELF:} embedded localisation of
  features in pre-trained {CNN},'' in \emph{ICCV}, 2019.

\bibitem{Simonyan14a}
K.~Simonyan, A.~Vedaldi, and A.~Zisserman, ``Deep inside convolutional
  networks: Visualising image classification models and saliency maps,'' in
  \emph{ICLRW}, 2014.

\bibitem{DB15a}
J.~Springenberg, A.~Dosovitskiy, T.~Brox, and M.~Riedmiller, ``Striving for
  simplicity: The all convolutional net,'' in \emph{ICLRW}, 2015.

\bibitem{pointCloudSal_ICCV19}
T.~Zheng, C.~Chen, J.~Yuan, and K.~Ren, ``Pointcloud saliency maps,'' in
  \emph{ICCV}, 2019.

\bibitem{RobotCarDatasetIJRR}
W.~Maddern, G.~Pascoe, C.~Linegar, and P.~Newman, ``{1 Year, 1000km: The Oxford
  RobotCar Dataset},'' \emph{IJRR}, vol.~36, no.~1, pp. 3--15, 2017.

\bibitem{kitti2013IJRR}
A.~Geiger, P.~Lenz, C.~Stiller, and R.~Urtasun, ``Vision meets robotics: The
  {KITTI} dataset,'' \emph{IJRR}, vol.~32, no.~11, pp. 1231 -- 1237, 2013.

\bibitem{3dfeatnet}
Z.~J. Yew and G.~H. Lee, ``3dfeat-net: Weakly supervised local 3d features for
  point cloud registration,'' in \emph{ECCV}, 2018.

\bibitem{li2019usip}
J.~Li and G.~H. Lee, ``Usip: Unsupervised stable interest point detection from
  3d point clouds,'' in \emph{ICCV}, 2019.

\bibitem{Georgakis2018}
G.~{Georgakis}, S.~{Karanam}, Z.~{Wu}, J.~{Ernst}, and J.~{Košecká},
  ``End-to-end learning of keypoint detector and descriptor for pose invariant
  3d matching,'' in \emph{CVPR}, 2018.

\bibitem{deepmindNIPS19}
T.~Kulkarni, A.~Gupta, C.~Ionescu, S.~Borgeaud, M.~Reynolds, A.~Zisserman, and
  V.~Mnih, ``Unsupervised learning of object keypoints for perception and
  control,'' in \emph{NeurIPS}, 2019.

\bibitem{VerdieYFL14}
Y.~Verdie, K.~M. Yi, P.~Fua, and V.~Lepetit, ``{TILDE:} {A} temporally
  invariant learned detector,'' in \emph{CVPR}, 2014.

\bibitem{lf_net}
Y.~Ono, E.~Trulls, P.~Fua, and K.~M. Yi, ``Lf-net: Learning local features from
  images,'' in \emph{NeurIPS}, 2018.

\bibitem{detone2018superpoint}
D.~DeTone, T.~Malisiewicz, and A.~Rabinovich, ``Superpoint: Self-supervised
  interest point detection and description,'' in \emph{CVPR}, 2018.

\bibitem{christiansen2019unsuperpoint}
P.~H. Christiansen, M.~F. Kragh, Y.~Brodskiy, and H.~Karstoft,
  ``{UnsuperPoint:} end-to-end unsupervised interest point detector and
  descriptor,'' \emph{arXiv preprint arXiv:1907.04011}, 2019.

\bibitem{cieslewski2019sips}
T.~Cieslewski, K.~G. Derpanis, and D.~Scaramuzza, ``{SIPs:} succinct interest
  points from unsupervised inlierness probability learning,'' in
  \emph{3DV}.\hskip 1em plus 0.5em minus 0.4em\relax IEEE, 2019, pp. 604--613.

\bibitem{truong2019glampoints}
P.~Truong, S.~Apostolopoulos, A.~Mosinska, S.~Stucky, C.~Ciller, and S.~D.
  Zanet, ``{GLAMpoints:} greedily learned accurate match points,'' in
  \emph{ICCV}, 2019, pp. 10\,732--10\,741.

\bibitem{tyszkiewicz2020disk}
M.~Tyszkiewicz, P.~Fua, and E.~Trulls, ``{DISK:} learning local features with
  policy gradient,'' \emph{NeurIPS}, vol.~33, 2020.

\bibitem{qi2016pointnet}
C.~R. Qi, H.~Su, K.~Mo, and L.~J. Guibas, ``{PointNet: Deep Learning on Point
  Sets for 3D Classification and Segmentation},'' in \emph{CVPR}, 2017.

\bibitem{qi2017pointnetplusplus}
C.~R. Qi, L.~Yi, H.~Su, and L.~J. Guibas, ``{PointNet++: Deep Hierarchical
  Feature Learning on Point Sets in a Metric Space},'' in \emph{NeurIPS}, 2017.

\bibitem{pointCNN_NIPS18}
Y.~Li, R.~Bu, M.~Sun, W.~Wu, X.~Di, and B.~Chen, ``Pointcnn: Convolution on
  x-transformed points,'' in \emph{NeurIPS}, 2018.

\bibitem{uypointnetvlad}
M.~A. Uy and G.~H. Lee, ``{PointNetVLAD: Deep Point Cloud Based Retrieval for
  Large-Scale Place Recognition},'' in \emph{CVPR}, 2018.

\bibitem{Xiang_2019_CVPR}
C.~Xiang, C.~R. Qi, and B.~Li, ``Generating {3D} adversarial point clouds,'' in
  \emph{CVPR}, June 2019.

\bibitem{pointFusion_CVPR18}
D.~Xu, D.~Anguelov, and A.~Jain, ``{PointFusion}: Deep sensor fusion for {3D}
  bounding box estimation,'' in \emph{CVPR}, 2018.

\bibitem{dh3d}
J.~Du, R.~Wang, and D.~Cremers, ``{DH3D:} deep hierarchical {3D} descriptors
  for robust large-scale {6DoF} relocalization,'' in \emph{ECCV}.\hskip 1em
  plus 0.5em minus 0.4em\relax Springer, 2020, pp. 744--762.

\bibitem{d3feat}
X.~Bai, Z.~Luo, L.~Zhou, H.~Fu, L.~Quan, and C.-L. Tai, ``{D3Feat:} joint
  learning of dense detection and description of {3D} local features,'' in
  \emph{CVPR}, 2020, pp. 6359--6367.

\bibitem{Selvaraju_2017_ICCV}
R.~R. Selvaraju, M.~Cogswell, A.~Das, R.~Vedantam, D.~Parikh, and D.~Batra,
  ``Grad-cam: Visual explanations from deep networks via gradient-based
  localization,'' in \emph{ICCV}, Oct 2017.

\bibitem{NIPS2018_8160}
J.~Adebayo, J.~Gilmer, M.~Muelly, I.~Goodfellow, M.~Hardt, and B.~Kim, ``Sanity
  checks for saliency maps,'' in \emph{NeurIPS}, 2018.

\bibitem{BaehrensJMLR10}
D.~Baehrens, T.~Schroeter, S.~Harmeling, M.~Kawanabe, K.~Hansen, and K.-R.
  M\"{u}ller, ``How to explain individual classification decisions,'' \emph{J.
  Mach. Learn. Res.}, vol.~11, pp. 1803--1831, Aug. 2010.

\bibitem{Kapur1985}
J.~Kapur, P.~Sahoo, and A.~Wong, ``A new method for gray-level picture
  thresholding using the entropy of the histogram,'' \emph{Computer Vision,
  Graphics, and Image Processing}, vol.~29, no.~3, pp. 273 -- 285, 1985.

\bibitem{ransac}
M.~A. Fischler and R.~C. Bolles, ``Random sample consensus: A paradigm for
  model fitting with applications to image analysis and automated
  cartography,'' \emph{CACM}, vol.~24, no.~6, pp. 381--395, 1981.

\end{thebibliography}

\end{document}